\documentclass[letterpaper, 10 pt, conference]{ieeeconf}

\IEEEoverridecommandlockouts                              
\usepackage{times}
\usepackage{multicol}
\usepackage{url}
\usepackage{color,soul}

\usepackage[dvipsnames]{xcolor}
\usepackage{subcaption}
\usepackage{amsmath,amssymb}

\usepackage{amsthm}
\usepackage{graphicx}  
\usepackage{tabularx, booktabs}

\usepackage[colorinlistoftodos]{todonotes}
\newcolumntype{C}{>{\centering\arraybackslash}X} 

\newtheoremstyle{myplain}
    {}
	{}
	{\itshape}
	{}
	{\bfseries}
	{}
	{5pt plus 1pt minus 1pt}
	{}

\newtheoremstyle{mydefinition}
	{}
	{}
	{\normalfont}
	{}
	{\bfseries}
	{}
	{5pt plus 1pt minus 1pt}
	{}
	
\theoremstyle{myplain}

\newcommand{\ourmethod}{RAMP}

\let\labelindent\relax
\usepackage{enumitem}

\usepackage[bookmarks=true]{hyperref}
\hypersetup{
    colorlinks=true,
    linkcolor=black,
    filecolor=black,
    urlcolor=magenta,
    pdftitle={RAMP: Hierarchical Reactive Motion Planning for Manipulation Tasks Using Implicit Signed Distance Functions},
}

\title{\LARGE \bf
RAMP: Hierarchical Reactive Motion Planning for Manipulation Tasks \\ Using Implicit Signed Distance Functions
}

\author{Vasileios Vasilopoulos, Suveer Garg, Pedro Piacenza, Jinwook Huh and Volkan Isler
\thanks{All authors are with the Samsung AI Center NY, 837 Washington St, New York, NY 10014. (e-mail: \texttt{$\{$vasileios.v, suveer.garg, p.piacenza, jinwook.huh, ibrahim.i$\}$@samsung.com}).}%
}

\renewcommand{\baselinestretch}{.96}

\begin{document}

\maketitle
\thispagestyle{empty}
\pagestyle{empty}

\begin{abstract}
We introduce Reactive Action and Motion Planner (\ourmethod{}), which combines the strengths of sampling-based and reactive approaches for motion planning. In essence, \ourmethod{} is a hierarchical approach where a novel variant of a Model Predictive Path Integral (MPPI) controller is used to generate trajectories which are then followed asynchronously by a local vector field controller. We demonstrate, in the context of a table clearing application, that \ourmethod{} can rapidly find paths in the robot's configuration space, satisfy task and robot-specific constraints, and provide safety by reacting to static or dynamically moving obstacles. \ourmethod{} achieves superior performance through a number of key innovations: we use Signed Distance Function (SDF) representations directly from the robot configuration space, both for collision checking and reactive control. The use of SDFs allows for a smoother definition of collision cost when planning for a trajectory, and is critical in ensuring safety while following trajectories. In addition, we introduce a novel variant of MPPI which, combined with the safety guarantees of the vector field trajectory follower, performs incremental real-time global trajectory planning. Simulation results establish that our method can generate paths that are comparable to traditional and state-of-the-art approaches in terms of total trajectory length while being up to 30 times faster. Real-world experiments demonstrate the safety and effectiveness of our approach in challenging table clearing scenarios. Videos and code are available at: \url{https://samsunglabs.github.io/RAMP-project-page/}
\end{abstract}


\section{Introduction}



Despite numerous advances in motion planning, existing solutions for high degree-of-freedom robots 
struggle to provide fast and reliable solutions in unknown environments. 
In particular, the cost of collision checking can be prohibitive for real-time applications and there is no well-established trade-off between global path optimality and local reactivity to dynamic obstacles or unexpected conditions. 



Current planning approaches can be divided into two major categories: reactive and non-reactive. Traditional reactive schemes, such as artificial potential fields \cite{Khatib_APF_1986} or navigation functions \cite{Rimon_NavigationFunctions_1992}, generally provide fast updates and guaranteed safety against obstacles. However, their limitations due to local minima or numerical stability issues are well-documented \cite{LaValle_Planning_2006}. Additionally, purely reactive schemes typically need implicit representations of obstacles, which are not easy to obtain in high-dimensional configuration spaces.

On the other hand, various non-reactive, primarily sampling-based algorithms such as Probabilistic Road Map (PRM) or Rapidly-exploring Random Trees (RRT) \cite{LaValle_Planning_2006} have enjoyed wide adoption due to their general applicability. However, such approaches typically optimize for a trajectory using a full explicit map of the environment and might need to plan from scratch when that map changes, resulting in slow and inefficient implementations that cannot easily adapt to environments explored online. In addition, the sequential manner in which these algorithms expand during planning makes parallelization and GPU acceleration a difficult task.

Model Predictive Control (MPC) schemes provide a middle ground between pure reactive control and open-loop sampling-based planning: they can account for obstacles incrementally and quickly adjust the resulting trajectory. 
Moreover, MPC schemes that rely on forward simulation of control inputs, such as Model Predictive Path Integral (MPPI) control \cite{Williams_MPPI_2018}, are  parallelizable and can be implemented on a GPU. 
However, their proposed control trajectories can drastically change between timesteps, producing jerky inputs which require post-processing (e.g., control input spline fitting). Unlike traditional planning methods, MPC schemes simply encode task completion, safety or other configuration constraints as cost functions in the optimization problem, which does not necessarily guarantee their satisfaction by the resulting trajectory.

\begin{figure}[t]
\centering
\includegraphics[width=0.98\columnwidth]{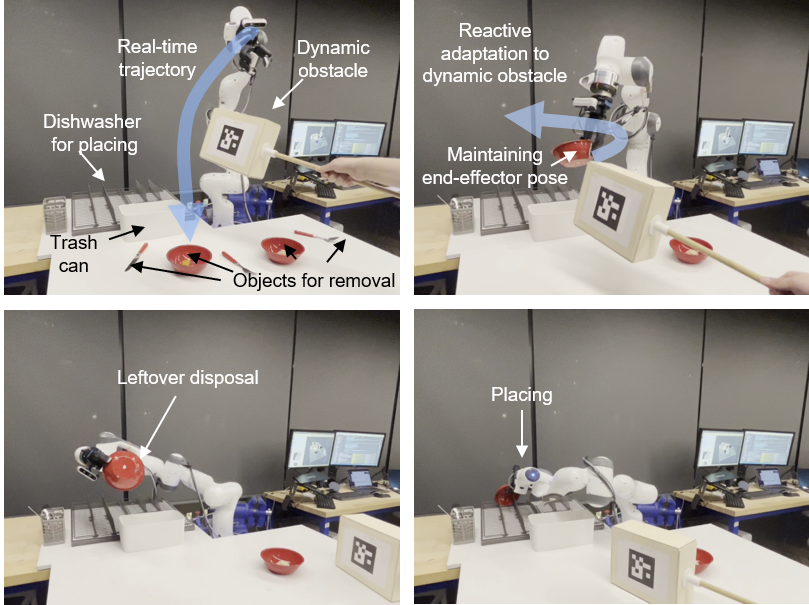}
\caption{{\bf Motivating scenario and execution example.} Motivated by table clearing scenarios that require picking, leftover disposal and placing, we develop a reactive motion planning architecture (\ourmethod) which can ensure safety against obstacles and provides fast, real-time trajectory updates, while respecting configuration and robot-specific constraints.}
\label{fig:motivating-scenario}
\end{figure}

The second major hurdle in designing fast motion planning algorithms is collision checking: 
each proposed configuration is checked for collisions with obstacles in the environment, which traditionally is a costly operation that requires the evaluation of several low-level geometric expressions \cite{PanChittaManocha_FCL_2012}. Alternative recent approaches estimate the probability of collision with neural networks and use it within an MPC algorithm \cite{DanielczukMousavianEppnerFox_ImplicitCollision_2021}. However, this does not necessarily ensure safety against obstacles in the environment. This problem is exacerbated when the task is to estimate distance to obstacles which is used when designing reactive algorithms.
Within the context of robotic manipulation, there is no algorithm that can take in robot configurations, output the distance of the robot to the closest workspace obstacle and its gradient, and use those values for fast, online reactive control.


\subsection{Outline of Contributions}

\begin{figure*}[t]
\centering
\includegraphics[width=0.98\textwidth]{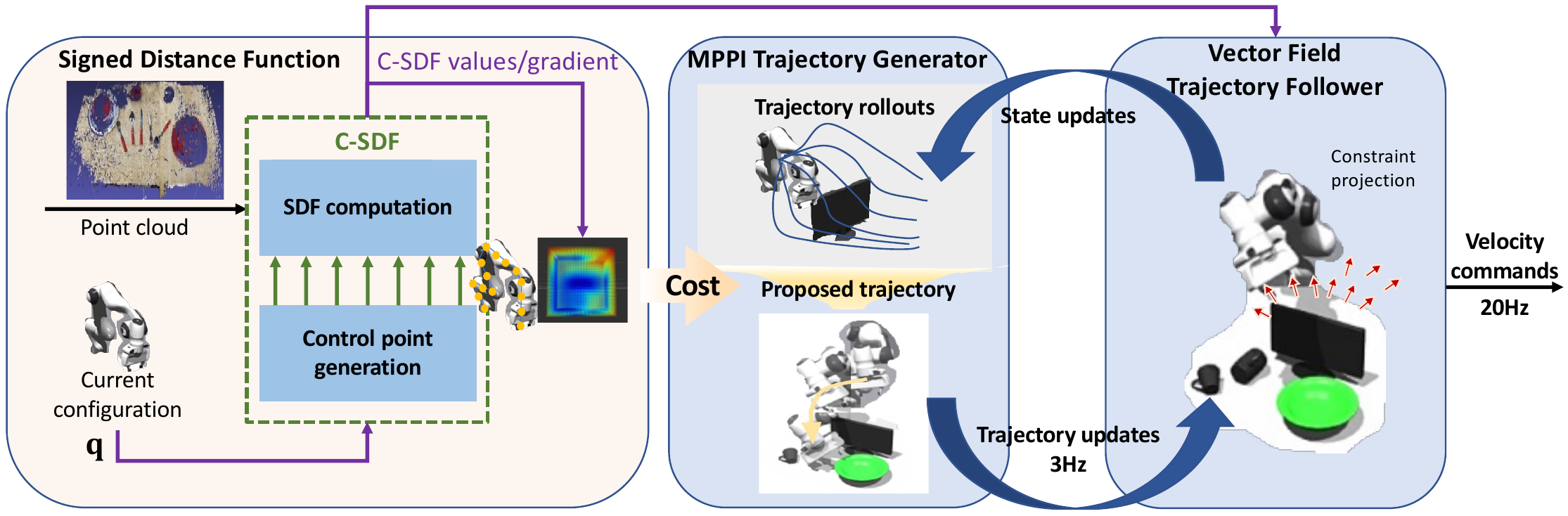}
\caption{\textbf{Overview of our online motion planning architecture (\ourmethod).} Assuming a known target joint configuration, the inputs to the system are the current joint configuration and the current point cloud scene representation, and the output is a desired velocity command. Our algorithm consists of two modules running in parallel and communicating asynchronously: an MPPI-based trajectory generator proposes trajectories to a vector field-based trajectory follower, which tracks the most recently proposed trajectory and avoids obstacles in real-time. Both modules use a Configuration Signed Distance Function (C-SDF) module, either for estimating collision costs of proposed configurations during planning (trajectory generator), or for avoiding obstacles by moving along the positive direction of the C-SDF gradient as needed (trajectory follower). Finally, the velocity commands from the trajectory follower are modified to handle any desired configuration-space constraints and passed to the low-level robot controller.}
\label{fig:architecture}
\end{figure*}

This work proposes Reactive Action and Motion Planner (\ourmethod), a hierarchical reactive scheme for high-dimensional robot manipulators (Fig.~\ref{fig:architecture}). In the proposed approach, a fast MPPI-based {\it trajectory generator} guides a local vector field-based {\it trajectory follower}, which generates real-time safe and smooth motions while satisfying constraints.

Our approach is the first one to use an implicit Signed Distance Function (SDF) module \cite{Park_DeepSDF_2019} for reactive control, computed directly using the robot's joint configuration. SDF representations are used both for fast collision checking for the trajectory generator and to represent the obstacles for the vector field construction during the trajectory following phase. The ability to parallelize direct distance queries enables our approach to run in real-time. 

The main contributions of our work are as follows:
\begin{itemize}
    \item {\it Configuration Signed Distance Function (C-SDF) module}: We present a module that takes in the robot's configuration and the scene's point cloud, and outputs an estimate of the distance of the entire robot body to nearby obstacles, along with its gradient. Unlike prior state-of-the-art methods in fast collision checking with neural networks \cite{DanielczukMousavianEppnerFox_ImplicitCollision_2021} which require offline training on many different scenes and only output a binary collision prediction, this module can be used either for MPC planning, as a proxy of configuration collision cost, or for online trajectory following, by using the C-SDF gradient to push the robot away from obstacles.
    \item {\it Online MPPI-based trajectory generator}: Unlike prior work in the literature, which uses MPPI as a low-level controller \cite{Bhardwaj_STORM_2022}, we configure MPPI as a fast, online, global trajectory generator, which takes in a start and a goal robot configuration, uses the C-SDF module for estimating configuration collision costs during planning and outputs a reference trajectory for the robot to track.
    \item {\it Online vector field-based trajectory follower}: We develop a closed-form, vector field-based module that tracks the proposed trajectory from the trajectory generator, uses the gradient of the C-SDF module for collision avoidance, and respects any provided configuration constraints (e.g., desired orientation angles of the end effector). This module sends smooth configuration-space velocity commands to the robot in real-time.
\end{itemize}

We demonstrate the speed and effectiveness of our approach quantitatively by comparing against a traditional planning algorithm and a state-of-the-art reactive control scheme. Additionally, we integrate our planning algorithm within a complete pipeline for table clearing tasks and qualitatively demonstrate its capabilities, both in static environments with multiple objects and in scenes with moving obstacles.

\subsection{Organization of the Paper}
The paper is organized as follows. Section~\ref{sec:related-work} discusses prior work on reactive motion planning. Section~\ref{sec:motivating-scenario} outlines the table clearing scenario that motivates our main motion planning task, described in Section~\ref{sec:problem-statement}. Section \ref{sec:architecture} offers an overview of our hierarchical architecture. Sections~\ref{sec:trajectory-generation} and \ref{sec:trajectory-following} describe our MPPI-based trajectory generator and vector field-based trajectory follower, respectively. Section~\ref{sec:numerical-experiments} provides our quantitative numerical results and comparison studies with other traditional and state-of-the-art methods, and Section~\ref{sec:real-world-experiments} describes our robot setup and includes qualitative real-world experiments. Finally, Section~\ref{sec:conclusion} concludes with a discussion and ideas for future work.

\section{Related Work}
\label{sec:related-work}

Since the original paper \cite{Williams_MPPI_2018}, Model Predictive Path Integral (MPPI) control has become quite popular in the reactive navigation literature. Recently, STORM \cite{Bhardwaj_STORM_2022} and SceneCollisionNet \cite{DanielczukMousavianEppnerFox_ImplicitCollision_2021} adapted it for reactive manipulation in high-dimensional spaces. These works consider static scenes and lack safety guarantees due to both the use of a neural approximator for collision checking and the absence of a safe low-level controller for trajectory following. Similar work \cite{Bowen_ClosedLoop_2014} performs closed-loop sampling-based global motion planning, by generating new collision-free plans based on a combination of a learned task model and adjusted Probabilistic Roadmaps (PRMs), but rely on expert demonstration. On the other hand, the authors in \cite{Schmitt_ReactiveManipulation_2019} construct safe, vector field-based controllers that encode mobile manipulation tasks, and sequence them using reinforcement learning. It has also been shown that vector field controllers can be used for online avoidance of singularities in robot manipulators \cite{Beck_SingularityAvoidance_2022} and safe following of a path provided by an external planner \cite{Arslan_ReferenceGovernors_2017}. Our work unifies MPPI-based planning approaches with vector field-based control for fast and safe online motion planning.

In terms of scene representations, in recent years, there has been a lot of interest in generating implicit representations of the workspace based on SDFs \cite{Park_DeepSDF_2019, Ortiz_iSDF_2022}. Such representations have already been used for offline path refinement (e.g., CHOMP \cite{Ratliff_CHOMP_2009}), learning models for task planning \cite{Driess_FunctionalsOfSDFs_2021}, predicting collision clearance in high-dimensional configuration spaces \cite{ChaseKew_NeuralCollisionClearance_2021} or, more recently, for querying the robot's distance to points in the scene during manipulation planning \cite{KoptevFigueroaBillard_NeuralImplicit_2023}. In our prior work \cite{Huh_C2GHOF_2021} we learn a neural representation of the global cost-to-go function and use its gradient to construct trajectories. This work generates the cost-to-go function over the entire workspace from an initial observation of the scene. There is no efficient way to update the cost-to-go function during execution.  In fact, at the moment there is no work in the literature that employs the gradients of workspace representations for online, reactive manipulation control. In our present work we set out to close this gap.




\section{A Motivating Scenario}
\label{sec:motivating-scenario}
The motion planning approach presented in this paper is motivated by a home robot application where a manipulator robot is charged with clearing a table by picking up objects from clutter, removing any leftover items by dropping them in a trash can, and then loading the objects in a dishwasher tray, without any human interventions (see Fig.~\ref{fig:motivating-scenario}). In this Section, we present an overview of this system which is also used for evaluating our algorithms later on.

We use the following object classes: plates, bowls, forks, spoons, and knives. The robot is equipped with a wrist camera for observing objects on the table, and our setup also includes an external camera for observing the whole scene and detecting other obstacles in the environment.

The perception module consists of off-the-shelf semantic segmentation \cite{KaimingGkioxari_MaskRCNN_2017} and object pose estimation \cite{He_FFB6D_2021} algorithms, trained on objects that the robot might encounter in the scene. Additionally, we use a custom grasping module, that proposes feasible grasps based on the detected object class, and a custom placement module, that proposes feasible leftover disposal and dishwasher placement poses. The robot needs to pick, dispose of leftover items and place objects sequentially in the dishwasher tray.

Our baseline motion planners are based on RRT \cite{LaValle_Planning_2006} or RRT* \cite{KaramanFrazzoli_RRTstar_2010}.
Some of their limitations, which we describe next, motivated the present work:
During system testing and evaluation, we observed that the robot sometimes needs to operate very close to its joint limits where sampling is problematic (e.g., when placing bowls in the dishwasher or when picking up plates from the table) and account for several constraints (e.g., the leftovers need to remain inside the plate before disposal). Typically, in such conditions, the baseline algorithms spend a lot of time planning (sometimes as long as 30 seconds) and, even when they do find a solution, a separate optimization module is still needed to design an open-loop joint-space trajectory that follows the generated path without violating any kinematic or dynamic constraints, adding to the total planning time. Additionally, it is not always straightforward to adjust online to changing obstacles in the scene, as the robot navigates to a given target.

\section{Problem Statement}
\label{sec:problem-statement}

We are now ready to formulate our motion planning problem. At each control timestep, given the current scene point cloud and robot configuration $\mathbf{q}$,  generate a velocity command $\mathbf{u}$ that drives the robot to the target $\mathbf{q}_{\text{goal}}$, while 
(i)~avoiding obstacles in the environment, (ii)~respecting any configuration constraints (such as maintaining a desired end effector pose), and (iii)~respecting the joint position and velocity limits of the robot.

Within our motivating scenario, the inputs to our system are obtained as follows: During the table clearing task, the motion planner is needed in order to navigate to a feasible end effector grasping, leftover disposal or dishwasher placement pose. We transform such a pose target to the robot's configuration space $\mathcal{C}$ using inverse kinematics \cite{Diankov_IKFast_2010}, to get an array of target joint angles, $\mathbf{q}_{\text{goal}} \in \mathcal{C}$.

For obstacle avoidance purposes, we consider the whole scene as a large point cloud. This point cloud is constructed by concatenating points sampled from meshes of known objects in the scene (e.g., the table or the dishwasher tray), with the point clouds of objects on the table (observed by the wrist camera) and external obstacles (observed by the external camera). Finally, assuming the presence of a low-level joint torque controller that can achieve commanded joint velocities $\mathbf{u} \in T\mathcal{C}$, we consider a first-order, fully-actuated abstraction of the robot for motion planning purposes, as $\dot{\mathbf{q}} = \mathbf{u}$.

\section{Architecture Overview}
\label{sec:architecture}

Our motion planning architecture is summarized in Fig.~\ref{fig:architecture}. It consists of two distinct modules running in parallel and communicating asynchronously: an MPPI-based {\it trajectory generator}, described in Section~\ref{sec:trajectory-generation}, and a vector field-based {\it trajectory follower}, described in Section~\ref{sec:trajectory-following}.

Given a target configuration $\mathbf{q}_{\text{goal}} \in \mathcal{C}$, the input to the trajectory generator is the current robot configuration $\mathbf{q} \in \mathcal{C}$ at each control iteration, and the output is a proposed trajectory in the configuration space, as a sequence of $N$ waypoints $\mathbf{q}_0, \ldots, \mathbf{q}_{N-1} \in \mathcal{C}$, connecting $\mathbf{q}$ and $\mathbf{q}_{\text{goal}}$ (i.e., $\mathbf{q}_0 = \mathbf{q}$ and $\mathbf{q}_{N-1} = \mathbf{q}_{\text{goal}}$). The trajectory follower tracks the most recent proposed trajectory from the trajectory generator and avoids obstacles in real-time, by generating velocity commands $\mathbf{u} \in T\mathcal{C}$ in the tangent space of the robot's configuration space. The trajectory generator asynchronously updates the proposed trajectory for the follower after each MPPI iteration.

Both modules use estimates of the robot body's signed distance to the scene point cloud given its joint configuration. We refer to these values as the {\it Configuration Signed Distance Function (C-SDF)} values. The trajectory generator uses C-SDF values to estimate the collision cost of proposed states during planning, whereas the trajectory follower uses the C-SDF value and gradient with respect to the current configuration in order to avoid obstacles. With GPU acceleration, we can afford to use an accurate C-SDF estimation algorithm, based on direct computation of distances between the robot and the scene. We describe this approach in Section~\ref{sec:c-sdf} next.
\section{Configuration Signed Distance Function}
\label{sec:c-sdf}

This Section describes our method for estimating the closest distance of the robot to obstacles in the environment, given its current configuration $\mathbf{q}$ and a point cloud representation of the scene. In Section \ref{subsec:control-points}, we describe the method for generating {\it control points} on the robot's body, which can be directly passed to existing Signed Distance Function (SDF) modules for distance estimation. Then, Section \ref{subsec:brute-force} describes our accurate and parallelized SDF representation, which we use both in the trajectory generator for performing batched collision cost queries of thousands of configurations in real-time, and in the trajectory follower for pushing away from obstacles using the SDF gradient. Finally, Section \ref{subsec:c-sdf} describes how we combine the SDF values for all the control points into a single Configuration SDF (C-SDF) value for each configuration.

\subsection{Control Point Generation}
\label{subsec:control-points}
Given a batch of $K \geq 1$ robot configurations $\{\mathbf{q}_i\}_{i=1,\ldots,K} \in \mathcal{C}$, the first step in rapidly estimating their C-SDF values is to generate {\it control points} that roughly represent the robot's placement in the workspace. To this end, we pick a set of $L$ {\it skeleton link frames} which coincide with some of the robot's joints, so that their pose in the workspace given a specific robot configuration can be easily computed using GPU-accelerated forward kinematics \cite{Sutanto_DifferentiableRobot_2020}. Then, we linearly interpolate between the locations of those $L$ frames to get a set of $C$ control points $\{\mathbf{c}_j(\mathbf{q}_i) \in \mathbb{R}^3 \}_{j=1,\ldots,C}$ for each configuration $\mathbf{q}_i$ (e.g., see inset of Fig.~\ref{fig:architecture}).

When the robot moves to a disposal or placement pose while grasping a particular object, we need to add points corresponding to the object itself to the overall list of control points, for accurate collision detection and distance estimation. To this end, assuming a known object geometry (in the form of a triangular mesh) and end effector pose during grasping, we sample points on the object's surface, transform them using forward kinematics and add them to the list of control points, for each configuration.

\subsection{Direct Computation of SDF Values}
\label{subsec:brute-force}
After the computation of the control points, we proceed to define an SDF representation in the robot's workspace. Namely, given a representation of the scene as a dense point cloud with $S$ points $\{\mathbf{s}_k \in \mathbb{R}^3\}_{k=1,\ldots,S}$, we can write the SDF value of a point $\mathbf{x} \in \mathbb{R}^3$ as 

\begin{equation}
\text{SDF}(\mathbf{x}) := \min_{k} ||\mathbf{x}-\mathbf{s}_k|| - \rho  
\end{equation}
with $\rho \in \mathbb{R}^+$ a small value that ensures the existence of both positive and negative SDF values and, therefore, the smoothness of the SDF zero level set. Even though this direct computation seems expensive, all distance queries between points $\mathbf{x} \in \mathbb{R}^3$ (which, in our case, coincide with the robot control points) and the point cloud can be easily parallelized on a GPU. In our experiments, we typically perform distance queries between hundreds of control points and thousands of points in the scene point cloud in less than 5ms.

\subsection{C-SDF Computation}
\label{subsec:c-sdf}
Given the above SDF representation, defined over the robot's workspace, and a configuration $\mathbf{q}$ with corresponding control points $\{\mathbf{c}_{j} (\mathbf{q})\}_{j=1,\ldots,C}$, we write the C-SDF of $\mathbf{q}$, conditioned on the $C$ control points as

\begin{equation}
\text{CSDF}_C(\mathbf{q}) := \min_j \text{SDF}\left((\mathbf{c}_j(\mathbf{q})\right) - r
\end{equation}
with $r \in \mathbb{R}^+$ a safety threshold. A positive C-SDF value implies that the robot is not in collision with the environment and vice versa.
\section{Trajectory Generation}
\label{sec:trajectory-generation}

\begin{figure}[t]
\centering
\includegraphics[width=0.98\columnwidth]{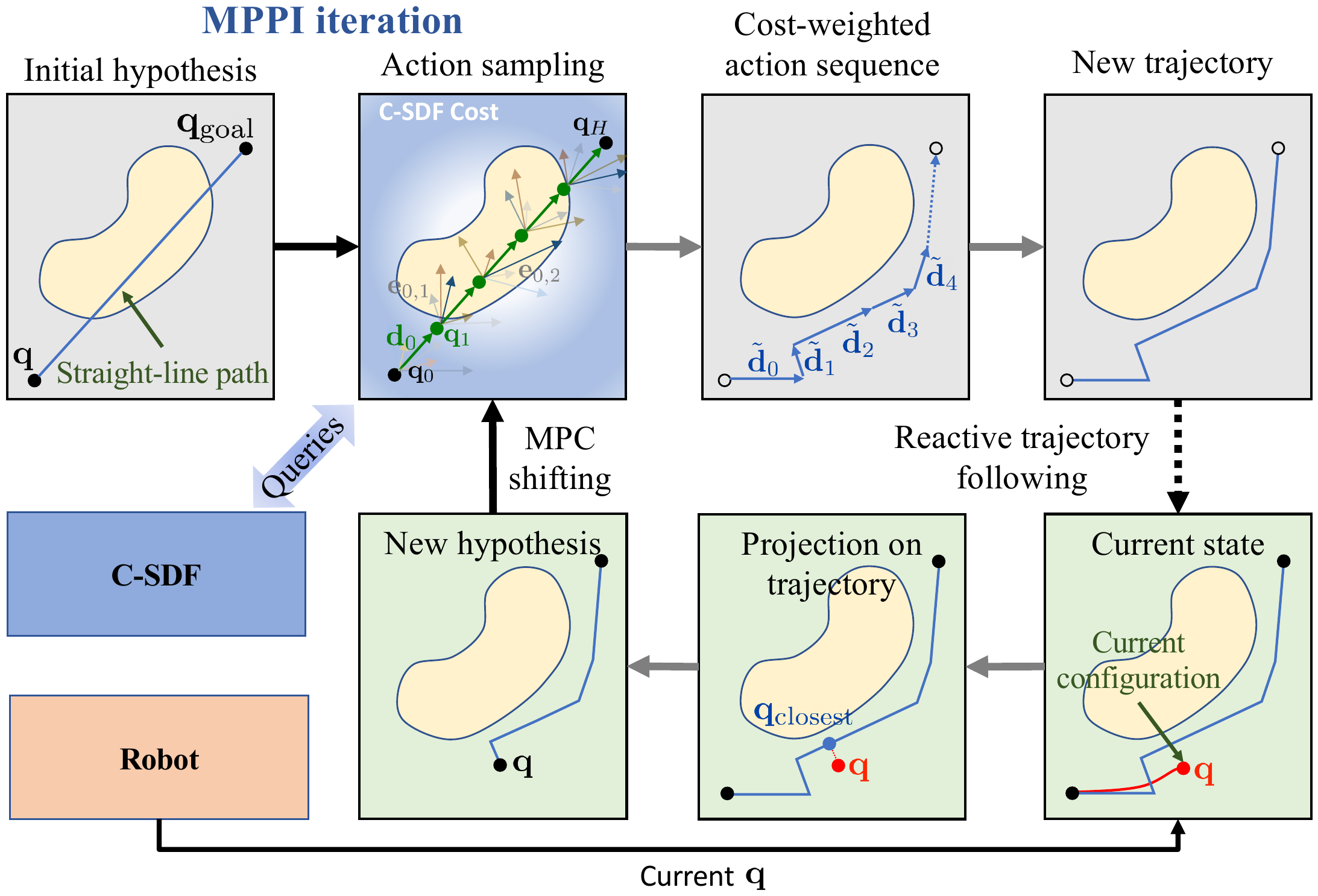}
\caption{Illustration of an MPPI trajectory planning iteration (top -- Section~\ref{subsec:model-sampling}), and our MPC ``shifting'' scheme (bottom -- Section~\ref{subsec:mpc-shifting}).}
\label{fig:mpc-updates}
\vspace{-2mm}
\end{figure}

This Section describes our MPPI-based global trajectory generation algorithm. MPPI is an MPC scheme that works by sampling multiple control sequences (control rollouts) around a nominal sequence and propagating them through the system's model. A new control sequence is evaluated as the weighted average of all rollouts (with weights computed according to some cost function) and is used to construct the nominal control sequence for the next iteration. Section~\ref{subsec:model-sampling} begins by describing our MPPI model and control sampling procedure. Then Section~\ref{subsec:cost-function} describes our cost function construction. Finally, Section~\ref{subsec:mpc-shifting} details our novel trajectory MPC shifting, necessitated by the asynchronous communication between the trajectory generator and the trajectory follower. We summarize our scheme in Fig.~\ref{fig:mpc-updates}.

\subsection{MPPI Adaptation to Trajectory Planning}
\label{subsec:model-sampling}
Our MPPI model is given as a discrete-time, continuous-state system $\mathbf{q}_{t+1} = \mathbf{q}_t + \mathbf{e}_t$, where $\mathbf{e}_t \sim \mathcal{N}(\mathbf{d_t}, \Sigma)$, with $\mathbf{d}_t$ a nominal (prior) displacement from $\mathbf{q}_t$ at time $t$ and $\Sigma$ its covariance. Unlike the original paper \cite{Williams_MPPI_2018} and other works \cite{Bhardwaj_STORM_2022} that use MPPI as a low-level (e.g., acceleration) controller, here we adapt the algorithm in order to perform trajectory planning, by using as our control input at time $t$ the displacement $\mathbf{d}_t$ from the configuration $\mathbf{q}_t$, and rely on our trajectory follower (Section~\ref{sec:trajectory-following}) for low-level control.

Hence, following \cite{Williams_MPPI_2018}, at each MPPI control iteration, we sample $M$ sequences of displacements $\mathbf{E}_j := \{\mathbf{e}_{j, 0}, \ldots, \mathbf{e}_{j, H-1}\}$, $j = 1, \ldots, M$, given a set of nominal configurations $\mathbf{q}_0, \ldots, \mathbf{q}_{H}$, and associated nominal displacements $\mathbf{d}_0, \ldots, \mathbf{d}_{H-1}$ for a control horizon $H$. We clamp the sampled displacements to ensure that they are within pre-defined magnitude limits and that they do not result in joint limit violations, and run them through the model to compute the associated rollout costs $C(\mathbf{E}_j)$. We then combine them by exponential averaging, to compute the posterior displacements after an MPPI iteration as
\begin{equation}
    \tilde{\mathbf{d}_t} = (1 - \alpha_\mathbf{d}) \, \mathbf{d}_t + \alpha_{\mathbf{d}} \, \frac{\sum_{j=0}^{M-1} w(\mathbf{E}_j) \, \mathbf{e}_{j,t}}{\sum_{j=0}^{M-1} w(\mathbf{E}_j)}
\end{equation}
with $\alpha_\mathbf{d} \in (0,1)$ a smoothing parameter, and
\begin{align*}
    & w(\mathbf{E}_j) := \frac{1}{\eta}\exp{\left(-\frac{1}{\lambda} \left( C(\mathbf{E}_j) + \lambda \sum_{t=0}^{H-1} \mathbf{d}_t^\top \, \Sigma^{-1} \, \mathbf{e}_{j,t} \right)\right)} \\
    & \eta := \sum_{j = 0}^{M-1} \exp{\left(-\frac{1}{\lambda} \left( C(\mathbf{E}_j) + \lambda \sum_{t=0}^{H-1} \mathbf{d}_t^\top \, \Sigma^{-1} \, \mathbf{e}_{j,t} \right)\right)}
\end{align*}
with $\lambda > 0$ a temperature parameter.

In our implementation, we initialize the MPPI loop with the hypothesis that the start and goal configurations are connected by a straight line path in joint space (see Fig.~\ref{fig:mpc-updates}). We discretize this path to find intermediate waypoints and the associated (nominal) displacements, which we use to start our MPPI updates. After each MPPI step, we update the trajectory tracked by the follower with the new configuration rollout $\{\tilde{\mathbf{q}}_t\}_{t = 0, \ldots, H}$, computed using the posterior displacements $\{\tilde{\mathbf{d}}_t\}_{t = 0, \ldots, H-1}$ (see top row of Fig.~\ref{fig:mpc-updates}). 

Since, as we describe next in Section~\ref{sec:trajectory-following}, the trajectory follower ensures safety against obstacles, we always append the target configuration to this updated trajectory, in order to bias the search toward the goal at the next MPPI iteration. Specifically, the updated trajectory forwarded to the trajectory follower is $\{\tilde{\mathbf{q}}_0, \ldots, \tilde{\mathbf{q}}_H, \mathbf{q}_{\text{goal}}\}$. Even if the line segment $\{\tilde{\mathbf{q}}_H, \mathbf{q}_{\text{goal}}\}$ is infeasible, our follower will repel against any obstacles in the environment, while waiting for an updated trajectory from the trajectory generator.

\subsection{Cost Function}
\label{subsec:cost-function}
Our cost function $C(\mathbf{E}_j)$ for each displacement rollout $\mathbf{E}_j$ is the sum of two terms; a running cost $C_r(\mathbf{E}_j)$ and a terminal cost $C_{\text{terminal}}(\mathbf{E}_j)$. Our running cost penalizes the total length of the trajectory, as well as collisions with the environment and self-collisions at each step of the horizon. Specifically, we write $C_r := C_{\text{length}} + C_{\text{coll}} + C_{\text{self-coll}}$, with
\begin{align}
    C_{\text{length}}(\mathbf{E}_j) & := w_{\text{length}} \, \sum_{t=0}^{H-1} ||\mathbf{e}_{j,t}|| \\
    C_{\text{coll}}(\mathbf{E}_j) & := w_{\text{coll}} \, \sum_{t=0}^{H-1} c_{\text{coll}}(\mathbf{q}_{j,t}, \mathbf{e}_{j,t}) \\
    C_{\text{self-coll}}(\mathbf{E}_j) & := w_{\text{self-coll}} \, \sum_{t=0}^{H-1} c_{\text{self-coll}}(\mathbf{q}_{j,t}, \mathbf{e}_{j,t})
\end{align}
with $w_{\text{length}}, w_{\text{coll}}, w_{\text{self-coll}} > 0$ tunable weights, and individual collision cost $c_{\text{coll}}$ that uses the C-SDF value of $\mathbf{q}_{j,t}$ as
\begin{equation}
  c_{\text{coll}}(\mathbf{q}_{j,t}, \mathbf{e}_{j,t}) :=
    \begin{cases}
      1 & \text{if CSDF$_C(\mathbf{q}_{j,t}) \leq \delta$} \\
      \frac{\delta}{\text{CSDF}_C(\mathbf{q}_{j,t})} & \text{if CSDF$_C(\mathbf{q}_{j,t}) > \delta$}
    \end{cases}       
\end{equation}
with $\delta$ a distance threshold. Finally, we estimate the individual cost of self-collision $c_{\text{self-coll}}(\mathbf{q}_{j,t}, \mathbf{e}_{j,t})$ using a neural network from \cite{Bhardwaj_STORM_2022} (MLP with 3 layers of [256,128,64] neurons and ReLU activations), trained on the robot kinematics.

To penalize distance from the goal $\mathbf{q}_{\text{goal}}$ for the last waypoint in the configuration rollout sequence, we define our terminal cost as
\begin{equation}
    C_{\text{terminal}}(\mathbf{E}_j) := w_{\text{terminal}} \, ||\mathbf{q}_{j, H} - \mathbf{q}_{\text{goal}}||
\end{equation}

\subsection{Trajectory MPC Shifting}
\label{subsec:mpc-shifting}
In traditional MPC schemes, the system executes the first $n$ steps in the control sequence and then re-plans. The optimization problem is warm-started by ``shifting'' the last computed control sequence by $n$. In our framework, that would imply stopping to re-plan after navigating to the $n$-th waypoint of the proposed trajectory rollout, which is similar to how prior work \cite{DanielczukMousavianEppnerFox_ImplicitCollision_2021} has implemented MPC-based trajectory planning. This approach results in non-smooth motions, with a lot of intermediate stops. 

On the contrary, in our framework, we let the trajectory follower track the last proposed trajectory, run MPPI asynchronously, and simply update the trajectory for the follower after each MPPI iteration. This necessitates a new scheme for MPC ``shifting'' to warm-start the next MPPI iteration, since it is not guaranteed that the robot will be exactly at the $n$-th waypoint of the followed trajectory after some time.

To this end, before starting the next MPPI iteration, we read the configuration state $\mathbf{q}$ of the robot and find the closest point $\mathbf{q}_{\text{closest}}$ to the previously proposed trajectory. We establish a new trajectory hypothesis for MPPI by discarding all waypoints that precede $\mathbf{q}_{\text{closest}}$, connecting $\mathbf{q}$ with $\mathbf{q}_{\text{closest}}$, and continuing the previously proposed trajectory from $\mathbf{q}_{\text{closest}}$ (see bottom row of Fig.~\ref{fig:mpc-updates}). 

We then proceed as described above, by discretizing this trajectory based on a desired distance threshold between nominal waypoints and establishing nominal displacements. It should be noted that this results in a variable MPPI horizon $H$ between different MPPI control iterations, which depends on the length of each trajectory hypothesis. This is another benefit of our modified MPPI scheme: intuitively, we need longer horizons and, therefore, more computation when the robot is far from $\mathbf{q}_{\text{goal}}$, and vice versa.
\section{Trajectory Following}
\label{sec:trajectory-following}

This Section describes our trajectory following scheme. Section~\ref{subsec:vector-field} details the construction of our velocity vector field, which guides the robot to the trajectory provided by the trajectory generator and ensures collision avoidance, and Section~\ref{subsec:constraints} describes our algorithm for handling constraints in real-time by modifying the vector field commands.

\subsection{Velocity Vector Field Construction}
\label{subsec:vector-field}
Given a configuration-space trajectory from the trajectory generator as a sequence of waypoints $\mathbf{q}_0, \ldots, \mathbf{q}_{N-1} \in \mathcal{C}$, the objective of the vector field-based follower is to generate joint velocity commands $\mathbf{u} \in T\mathcal{C}$, that track the provided trajectory while avoiding obstacles in the environment. To this end, each time the trajectory generator provides a new trajectory, we generate an arclength parameterization $\mathbf{r}(s): [0, 1] \rightarrow \mathcal{C}$, such that $\mathbf{r}(0) = \mathbf{q}_0$ and $\mathbf{r}(1) = \mathbf{q}_{N-1}$.

During execution, given the current robot configuration $\mathbf{q}$, we first compute its C-SDF value, $\text{CSDF}_C(\mathbf{q})$, as described in Sections~\ref{subsec:brute-force} and \ref{subsec:c-sdf}. By construction of the C-SDF, a sphere centered at any of the robot's control points at configuration $\mathbf{q}$ with radius equal to $\text{CSDF}_C(\mathbf{q}) > 0$ is collision-free. Hence, inspired by \cite{Arslan_ReferenceGovernors_2017}, we pick as target $\mathbf{r}(s^*)$ at configuration $\mathbf{q}$ the furthest point along the trajectory, with control points at most $\text{CSDF}_C(\mathbf{q})$ away from the control points at $\mathbf{q}$. This immediately implies that $\mathbf{r}(s^*)$ is also safe. Formally, we write
\begin{align*}
    s^* = s^*(\mathbf{q}) := & \max \left( \left\{ s \in [0,1] \, | \right. \right. \nonumber \\
    & \left. \left. || \mathbf{x}_j(\mathbf{r}(s)) - \mathbf{x}_j (\mathbf{q}) || \leq \text{CSDF}_C(\mathbf{q}), \, \forall
    j \right\} \right)
\end{align*}

Based on this target, we can define a potential field for trajectory following as
\begin{equation}
    \varphi(\mathbf{q}) := \frac{||\mathbf{q} - \mathbf{r}(s^*(\mathbf{q}))||^2 + \varepsilon}{\text{CSDF}_C(\mathbf{q}) + \varepsilon}
\end{equation}
with $\varepsilon \in \mathbb{R}^+$ a small value. Our velocity commands are then obtained from the vector field
\begin{equation}
    \mathbf{u}(\mathbf{q}) := -k \, \nabla \varphi(\mathbf{q})
\end{equation}
with $k \in \mathbb{R}^+$ a positive constant.

\subsection{Constraint Handling}
\label{subsec:constraints}
Within our setting of table clearing, sometimes our motion planning task has additional configuration constraints. For example, when disposing leftovers, the robot's end effector needs to maintain its relative orientation with the world $z$-axis, in order to keep the leftovers inside the object until disposal. To satisfy this particular constraint, for each object category, we identify an axis of the end effector frame whose relative angle with the world $z$-axis needs to be zero during movement to the disposal pose. Let $\hat{\mathbf{e}}(\mathbf{q})$ denote the unit vector representing this axis in the world frame, derived by the joint configuration $\mathbf{q}$ through forward kinematics, and $\hat{\mathbf{z}}$ denote the unit vector of the world $z$-axis. Then, formally, our constraint can be written as
\begin{equation}
\mathbf{h}(\mathbf{q}) := \left(\angle(\hat{\mathbf{e}}(\mathbf{q}), \hat{\mathbf{z}})\right) ^2 = \arccos^2(\hat{\mathbf{e}}(\mathbf{q}) \cdot \hat{\mathbf{z}}) = 0    
\end{equation}

Given this expression for $\mathbf{h}$ and a raw velocity command $\mathbf{u} \in T\mathcal{C}$, computed as described in Section \ref{subsec:vector-field}, we project $\mathbf{u}$ to the constraint's null space and add a gradient term to drive the deviation from the constraint manifold to zero. Specifically, we define our modified velocity command  $\mathbf{u}_\mathbf{h}$ that is passed to the low-level torque controller as

\begin{equation*}
    \mathbf{u}_\mathbf{h}(\mathbf{q}): = \left(I_{d \times d} - D\mathbf{h}^\top (D\mathbf{h} \, D\mathbf{h}^\top)^{-1} \, D\mathbf{h} \right) \mathbf{u} - c \, \nabla \mathbf{h}(\mathbf{q})
\end{equation*}
with $c \in \mathbb{R}^+$ a positive constant. We scale this command accordingly, to ensure that it does not violate the robot's joint velocity limits.
\section{Numerical Experiments}
\label{sec:numerical-experiments}

\begin{figure}[th]
\centering
\includegraphics[width=0.98\columnwidth]{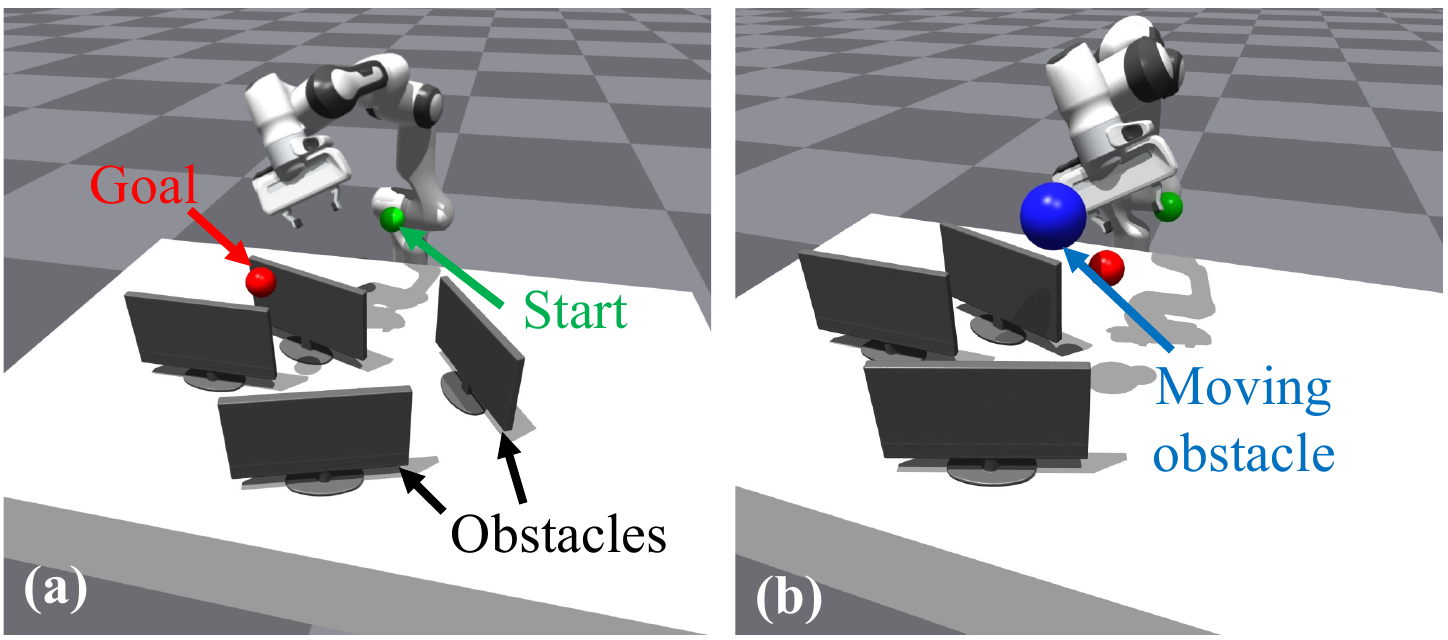}
\caption{Examples of generated simulation environments: (a) Static environment, (b) Environment with a moving obstacle.}
\label{fig:isaac-gym}
\vspace{-2mm}
\end{figure}

\begin{table}
\caption{Static Environment Comparison.}
\label{table:static-environment}
\begin{tabularx}{\columnwidth}{@{} lccc}
\toprule
& RRT$^*$ & SceneCollisionNet & \ourmethod \\
&  & \cite{DanielczukMousavianEppnerFox_ImplicitCollision_2021} & (Ours) \\
\midrule
Success rate $\uparrow$ & {\bf 1.0} & 0.86 & {\bf 1.0} \\
Planning time (sec) $\downarrow$ & 3.75$\pm$1.49 & 1.55$\pm$0.55 & {\bf 0.11$\pm$0.034} \\
Trajectory length $\downarrow$ & 1.0 & {\bf 0.91} & 1.04 \\
\bottomrule
\end{tabularx}
\end{table}

\begin{table}
\caption{Reactivity Comparison with a Moving Obstacle.}
\label{table:reactivity}
\begin{tabularx}{\columnwidth}{@{} lccc}
\toprule
& RRT$^*$ & SceneCollisionNet & \ourmethod \\
&  & \cite{DanielczukMousavianEppnerFox_ImplicitCollision_2021} & (Ours) \\
\midrule
Collision-free trials (\%) $\uparrow$ & 60 & 52 & \bf{94} \\
Safety (cm) $\uparrow$ &  2.5 $\pm$ 3.6 & 2.1 $\pm$ 3.2 & \bf{5.3$\pm$2.5}  \\
\bottomrule
\end{tabularx}
\end{table}

In this Section, we compare our method with traditional and state-of-the-art approaches for trajectory generation in static environments and environments with moving obstacles. For our baseline approaches, we choose the Rapidly-exploring Random Tree (RRT*) \cite{KaramanFrazzoli_RRTstar_2010} algorithm in order to compare path lengths against an asymptotically-optimal well-established method, and SceneCollisionNet \cite{DanielczukMousavianEppnerFox_ImplicitCollision_2021} as a representative state-of-the-art approach that also uses MPPI for trajectory planning. Since both RRT* and SceneCollisionNet rely on a collision checking algorithm, we use the Flexible Collision Library (FCL) for both. We apply RRT* using the Open Motion Planning Library (OMPL) \cite{sucan2012the-open-motion-planning-library}, and allow it 5 seconds for trajectory optimization with 5\% goal-biased sampling for better performance. Finally, for a more fair comparison of overall path length with methods that do not include a reactive trajectory follower, we just use the trajectory generation part of our algorithm (without the reactive follower) in static environments, and we only include the reactive follower in dynamic environments.

For our quantitative comparisons in static environments, we use Isaac Gym \cite{liang2018gpu} and evaluate performance in 5 different cluttered scenes (see Fig.~\ref{fig:isaac-gym}a). For dynamic environments, we set up the same cluttered environments and add a single moving obstacle traversing the scene with constant velocity (see Fig.~\ref{fig:isaac-gym}b). For each setting, we generate 100 random start and goal pair configurations. The start and goal configurations are sufficiently far from each other and have end effector locations close to the table in order to make the planning problem more difficult.

\subsection{Comparison in Static Environments}

In static environments, we measure overall performance in terms of success rate, planning time, and trajectory length. The trajectory length is normalized by the RRT* trajectory length. Table \ref{table:static-environment} shows the comparison results with the used metrics in static environments. For all methods, we average planning times and lengths of successfully converged trajectories. SceneCollisionNet has an 86\% success rate, compared to our 100\% success rate. This can be partly attributed to the fact that, unlike our global trajectory generation, SceneCollisionNet simply perturbs the greedy direction to the goal configuration at each MPPI planning step, and then removes rollouts that collide with obstacles, which could result in sometimes getting stuck behind obstacles. Additionally, our method is up to 14 times faster on average compared to SceneCollisionNet and up to 34 times faster on average compared to RRT*. On the other hand, our method generates 4\% longer trajectories on average compared to RRT*. This occurs because our MPPI planning procedure focuses on fast feasible trajectory generation and we rely on the trajectory follower to produce a smooth trajectory in real-time. On the contrary, SceneCollisionNet's trajectories comprise of piecewise straight lines, which help in reducing the total path length. Overall, the results establish that our algorithm can generate trajectories with high success rates and in a faster way than other state-of-the-art approaches.

\subsection{Comparison in Dynamic Environments}
In dynamic environments with a moving obstacle, we evaluate the performance of each method in terms of percentage of collision-free iterations, and a safety metric. We define the safety metric as the average minimum distance between the robot and obstacles (both dynamic and static) throughout each planning trial in a scene. This metric gives us an insight regarding how close each method gets to the scene obstacles as they navigate to different targets. During a collision, the instantaneous value of this safety metric is zero. For these trials, in each scene, RRT* uses the first scene observation to generate a plan. On the other hand, SceneCollisionNet and our method receive periodic new observations of the scene.

Table \ref{table:reactivity} shows the comparison between the three methods for these metrics. SceneCollisionNet and RRT* have a lower number of collision-free iterations compared to our method. This can be attributed to our reactive trajectory follower module which helps us adapt to moving obstacles, unlike other methods that are not well equipped to deal with such conditions. Also, the results show that our method tends to maintain a higher distance to obstacles on average. This verifies that our method achieves safer motion in dynamic environments compared to alternatives.

\section{Real-World Evaluation}
\label{sec:real-world-experiments}

\begin{figure*}[t]
\centering
\includegraphics[width=0.95\textwidth]{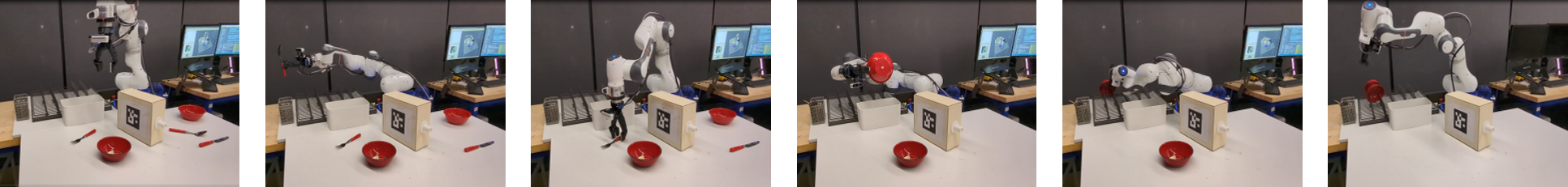}
\caption{{\bf Real-world experiment in a static environment.} We demonstrate the performance of \ourmethod{} in various static environments with bowls, plates, and utensils. The robot successfully performs table clearing tasks requiring collision-free navigation to specific end effector poses or joint configurations.}
\label{fig:experiment-static}
\vspace{-3mm}
\end{figure*}

\begin{figure*}[t]
\centering
\includegraphics[width=0.95\textwidth]{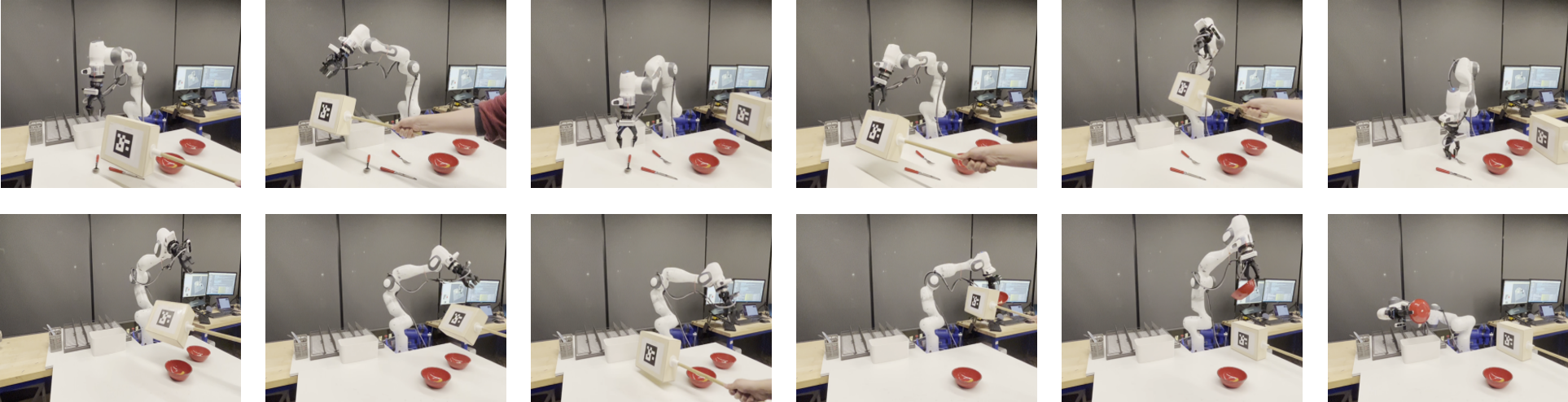}
\caption{{\bf Real-world experiment with a moving obstacle.} The robot approaches target objects while avoiding the moving obstacle or keeping a safe distance from it. During leftover disposal, it maintains the pose of the grasped object to prevent spills of contained items, while avoiding collisions.}
\label{fig:experiment-moving}
\vspace{-3mm}
\end{figure*}

\subsection{Experimental Setup}

We execute our experiments with a Franka Emika Panda robot, equipped with an Intel Realsense D435i wrist camera. We also use an Intel Realsense D455 camera that observes the scene and detects in real-time AprilTags that correspond to external obstacles of known geometry. The whole dishwasher loading pipeline that we describe in Section~\ref{sec:motivating-scenario} runs on an Ubuntu machine with 4 Nvidia RTX 2080 Ti GPUs and 12 Intel Core i9 CPU cores, and is implemented in ROS 2. Both our trajectory generator and our trajectory follower are implemented in PyTorch.
Our trajectory generator uses $M=500$ rollouts at each MPPI iteration, with an action sampling covariance $\Sigma = 0.005 \cdot I$ and $\lambda = 1.0$. We implement our C-SDF module with $\rho = 2$cm and $r = 5$cm. For collision checking, we use a distance threshold $\delta = 5$cm. Finally, the control gains of our trajectory follower are $k = c = 0.5$.

\subsection{Robot Experiments}

Similarly to our numerical experiments, in order to qualitatively verify our approach, we run our motion planning algorithm with the whole dishwasher loading pipeline in 5 static and 5 dynamic environments. For each scene, we randomly generate tabletop scenarios with a maximum number of 5 objects to be loaded to the dishwasher tray. In static scenes, we place a box-shaped obstacle among the tabletop objects to make the problem more difficult (see Fig.~\ref{fig:experiment-static}) whereas in dynamic scenes, a human operator randomly moves the same box-shaped obstacle to perturb the execution of the task (see Fig.~\ref{fig:experiment-moving}). The reader is also referred to the accompanying video submission for examples.

The robot managed to successfully place 19/21 objects totally in the static environments, with one grasping failure and one collision. Of those 19 objects, 7 also required leftover disposal before placing. The grasping failure was not directly related to our motion planning algorithm, since it was associated with a lack of a feasible inverse kinematics solution when grasping one of the plates, whereas the collision was related to the asymmetric shape of link 5 of the Panda robot, which cannot be entirely captured by control points on the body skeleton. We believe that sampling control points directly from the meshes of the robot links can resolve this issue, but might make the algorithm slightly slower.

In the dynamic scene scenarios, the robot managed to place all 18 objects in total. Of those 18 objects, 8 required leftover disposal before placing. We recorded two minor collisions with the obstacle box: one in which we moved the box too fast toward the robot (which we consider an adversarial case), and one in which the box was briefly behind the robot body and was not visible by the external camera observing the scene. Our results and success rates demonstrate the effectiveness and reliability of our reactive motion planning architecture, in complex real-world scenarios with non-trivial placement tasks.
\section{Conclusion}
\vspace{-1mm}
This paper presented a reactive motion planning architecture (\ourmethod{}) which integrates a fast MPPI-based trajectory generator and an online vector field-based trajectory follower. To ensure safety while trajectory following in dynamic environments with  obstacles, \ourmethod{} uses Signed Distance Function (SDF) representations, both for fast collision checking and real-time reactive control. 
Our method can generate collision-free and smooth trajectories and execute them while reacting to dynamic obstacles.
Experiments indicate that \ourmethod{} is significantly faster than the baselines and ensures safety by minimizing collisions, even with moving obstacles.
In table clearing scenarios, 
\ourmethod{} successfully picked up, disposed leftovers from and placed 37 out of 39 objects.
In our future work, we would like to incorporate multi-modal sensors (e.g., tactile or acoustic sensors) for feedback, and further investigate online implicit scene representations.


\label{sec:conclusion}

\small
\bibliographystyle{IEEEtran}
\bibliography{references}

\end{document}